\def\year{2018}\relax
\documentclass[letterpaper]{article} 
\usepackage{aaai18}  
\usepackage{times}  
\usepackage{helvet}  
\usepackage{courier}  
\usepackage{url}  
\usepackage{graphicx}  
\usepackage{amsmath,amssymb}
\usepackage{enumitem}
\usepackage{tikz} 
\usepackage{adjustbox}
\usepackage{mathtools}
\DeclareMathOperator{\E}{\mathbb{E}}
\DeclarePairedDelimiter\norm{\lVert}{\rVert}

\usepackage{fancyhdr}
\usepackage{lipsum}
\usepackage{etoolbox}

\patchcmd{\maketitle}
  {\end{titlepage}}
  {\thispagestyle{titlepagestyle}\end{titlepage}}
  {}{}

\fancypagestyle{titlepagestyle}
{
   \fancyhf{}
   \fancyhead[C]{$^{*}$Published as a conference paper at Computing in Cardiology (CinC),  2017.}
   
}

\begin{document}
\title{Learning disentangled representation from 12-lead electrograms:  \\application in localizing the origin of Ventricular Tachycardia}
\author{Prashnna K Gyawali,$^{1}$ B. Milan Horacek,$^2$ John L. Sapp,$^2$ Linwei Wang$^{1}$\\
$^{1}$ Rochester Institute of Technology, Rochester, USA\\
$^{2}$ Dalhousie University, Halifax, NS, Canada\\
$\left\{pkg2182@rit.edu\right\}$
}

\maketitle
\begin{abstract}
The increasing availability of electrocardiogram (ECG) data has motivated the use of data-driven models for automating various clinical tasks based on ECG data. The development of subject-specific models are limited by the cost and difficulty of obtaining sufficient training data for each individual. The alternative of population model, however, faces challenges caused by the significant inter-subject variations within the ECG data. We address this challenge by investigating for the first time the problem of learning representations for clinically-informative variables while disentangling other factors of variations within the ECG data. In this work, we present a conditional variational autoencoder (VAE) to extract the subject-specific adjustment to the ECG data, conditioned on task-specific representations learned from a deterministic encoder. To encourage the representation for inter-subject variations to be independent from the task-specific representation, maximum mean discrepancy is used to match all the moments between the distributions learned by the VAE conditioning on the code from the deterministic encoder. The learning of the task-specific representation is regularized by a weak supervision in the form of contrastive regularization. We apply the proposed method to a novel yet important clinical task of classifying the origin of ventricular tachycardia (VT) into pre-defined segments, demonstrating the efficacy of the proposed method against the standard VAE.  
\end{abstract}

\section{Introduction}
The increasing availability of electrocardiogram (ECG) data has motivated the use of data-driven models for automating various clinical tasks based on ECG data \cite{DBLP:journals/corr/RajpurkarHHBN17}. However, ECG signals exhibit significant physiological variations across individuals \cite{park2012using}, \cite{DBLP:journals/corr/RajpurkarHHBN17}, \cite{wiens2010active}, a manifestation of anatomical differences in the heart and torso among individuals along with electrode positioning on the body surface during ECG acquisition \cite{plonsey}. There are also various pathological confounders that are tied to the specific clinical task under consideration, such as sub-groups under a population group, or different characteristics of structural abnormality of the heart. These inter-subject variations are not always observed and they present a great challenge for building accurate population-based models for patient-specific uses. At the same time, the alternative of patient-specific models are difficult to implement in clinical practice due to the challenge and cost in obtaining sufficient data from each patient.

\begin{figure}[tb]
\begin{center}
  \includegraphics[width=0.47\textwidth]{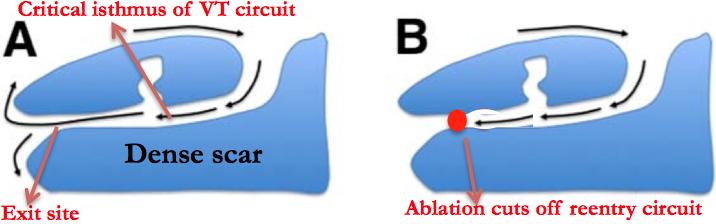}
\end{center}
\caption{\small{Schematics of a VT reentry circuit. \textbf{A}: an electrical "short circuit" travels through narrow strands of
surviving tissue inside the scar and exits from
the scar to depolarize the rest of the ventricle. \textbf{B}: Cutting off the isthmus or exits that form the "short circuit" in catheter ablation. }} 
\label{fig:VTcircuit}
\end{figure}

One particular example is the clinical task of localizing the origin of ventricular tachycardia (VT) from an ECG signal. VT is a significant cause of sudden cardiac deaths in the United States \cite{stevenson2009ventricular}. Such VT usually involves an electrical ``short circuit" formed by narrow strands of surviving tissue inside the myocardial scar as illustrated in Fig. \ref{fig:VTcircuit}A. Clinically, this is treated by cutting off the circuit at its exit (VT exit) from the scar using catheter ablation \cite{stevenson2009ventricular} as shown in Fig. \ref{fig:VTcircuit}B. Because a VT exit acts like the origin of ventricular activation to be seen on ECG, these exit sites are currently located by repeatedly stimulating (i.e. pacing) different locations of the heart until localizing the site where pacing reproduce the QRS morphology of the VT on all 12 leads of the ECG \cite{stevenson2009ventricular}.

Rather than this ``trial-and-error" status quo, naturally, an automatic real-time identification of the VT exit from 12-lead ECG data would improve the efficiency of ablation procedure. Limited attempts by the use of methods such as template matching \cite{sapp2012automated} and support vector machines \cite{yokokawa2012automated} have been made to use the data from a large cohort of patient to learn the relationship between the origin of ventricular activation and the ECG data. However, limited accuracy has been reported so far. One of the major challenges, as mentioned earlier, is the significant inter-subject variations across individuals \cite{park2012using}.

In this paper, we investigate for the first time the use of deep generative models \cite{kingma2013auto} \cite{DBLP:conf/icml/RezendeMW14} to explicitly separate subject-specific variations in the ECG data from the hidden representation informative for the clinical task at hand. Learning to disentangle the underlying causal factors of the data has received increasing attention in the recent advancement of deep learning \cite{bengio2013representation}. A variety of models, many built off variational autoencoder (VAEs) and generative adversarial network (GANs), have been shown to be effective in learning deep representations informative for the task at hand while removing the non-informative factors of variations in domains such as computer vision \cite{mathieu2016disentangling},  computer graphics \cite{kulkarni2015deep} and natural language processing \cite{hu2017toward}. Outside these domains and into domain such as medicine and biology, however, the notion of learning disentangled representations to improve the task at hand has been little explored.

In this paper, we recognize the challenge of inter-subject variations inside clinical data with the potential solution of learning disentangled representations. In specific, we deal with a scenario where training labels for the task-specific representations are available and none of the factors of variations is observed. To achieve disentanglement in this setting, we leverage both variational autoencoders (VAEs) \cite{kingma2013auto} \cite{DBLP:conf/icml/RezendeMW14} and maximum mean discrepancy (MMD) \cite{gretton2007kernel} along with contrastive regularization. Formally, for the specific task of predicting VT origin from ECG data, we aim to learn a latent representation $\mathbf{z}$ to capture the individual level physiological and pathological variations,  and a representation for the task-specific factor (VT factor) $\mathbf{v}$ to capture the relationship between origins of ventricular activation and ECG data as defined by the underlying biophysics. The latter is learned by a deterministic encoder, while the former is learned by a variational encoder conditioned on the latter. These two representations are encouraged to be statistically independent by matching all order of moments of the marginal posterior distribution $q(\mathbf{z}|\mathbf{v}_{i})$ with $q(\mathbf{z}|\mathbf{v}_{j})$ where $i \neq j$ and $k = \{1,..,i,..,j,..,c\}$ represents $c$ categories of the task-specific representation $\mathbf{v}$. To further increase the discriminative power of the task-specific representation, contrastive regularization is imposed on the deterministic encoder for a weak supervision given the label information in the training data. Our main contributions include:
\begin{itemize}
\item We investigate the first use of deep generative models to disentangle inter-subject variations within ECG data from the representation informative for clinical task at hand. 
\item We propose a variational encoder conditioned on the codes of a deterministic encoder, to learn to separate inter-subject variations from task-specific representations while learning the latter with a weak supervision via contrastive loss.
\item We propose a mechanism to encourage the statistical independence between the task-specific representation and inter-subject variations by matching all order of moments between the distributions learned from conditional VAE for the latter conditioning on the former.
\item We demonstrate the performance of our approach on a challenging clinical task of localizing the origin of VT from ECG data, and compare its performance with the standard VAE. 
\end{itemize}

The article is organized as follows. We introduce several related works in the next section. We provide a brief background of some preliminaries about variational autoencoder and maximum mean discrepancy. In model section, we present our proposed methodology along with training procedure. In experiment section, we present qualitative and quantitative analysis of presented model. Finally, we conclude the paper with limitations and possible future avenue.

\section{Related Works}
It has been widely accepted that different representations can entangle and hide explanatory factors of variation to a different extent within the data \cite{bengio2013representation}. The challenge of separating factors of variation from the data has been a topic of emerging interest in computer vision and related fields. As an early approach, a bilinear model \cite{tenenbaum2000separating} was proposed to capture sufficiently expressive representations of factors of variations in the data. They used the proposed model to separate the handwriting style from the content when recognizing handwritten digits.
The work in \cite{chen2017} presents the discriminative architecture using denoising autoencoder to address inter-subject variations in ECG data.
In the context of generative models,  the work in \cite{desjardins2012disentangling} and \cite{reed2014learning}
used restricted Boltzman machine to model multiplicative interactions between latent factors of variation, 
separating identity from emotion \cite{desjardins2012disentangling} 
or expression from pose \cite{reed2014learning} in facial images.

Recent works in learning disentangled representation started to exploit the power of deep generative models like VAE and generative adversarial networks (GAN). The work in \cite{mathieu2016disentangling} performs unsupervised disentanglement leveraging GAN. Similarly, the work in \cite{chen2016infogan} used information-theuritic approach to learn disentangled representations in GAN framework. However, training such adversarial network requires careful optimization of a difficult minimax problem \cite{li2015generative}. 

The work in \cite{higgins2016early} used a VAE framework where they demonstrated that by enforcing independence of features in the latent state it is possible to learn representations generalizing well to the new task. In \cite{kingma2014semi}, a conditional VAE was used to demonstrate the disentanglement between the label information and the style in a semi-supervised setting by providing an additional one-hot vector as an input to the generative model. The work in \cite{DBLP:journals/corr/LouizosSLWZ15} extended this idea by adding MMD to further encourage statistical independence.  The work in both \cite{kingma2014semi} and \cite{DBLP:journals/corr/LouizosSLWZ15}, however, requires a certain representations to be observed by conditioning the other representations on it. The work in \cite{kulkarni2015deep} explored the disentanglement of content and style in the domain of computer graphics and produced impressive result benefiting from unique training procedure requiring strong supervision in designing the data.

\section{Background}
\subsection{Variational Autoencoder}
The variational autoencoder (VAE) \cite{kingma2013auto}, \cite{DBLP:conf/icml/RezendeMW14} is a generative model where the data $\mathbf{X}$ is generated by the generative distribution $p_{\theta}\mathbf{(X|z)}$ involving a set of latent variables $\mathbf{z}$, which follows a prior distribution of $p\mathbf{(z)}$. This process can be defined as:
\begin{center}
$\mathbf{z}$ $\sim$ $p\mathbf{(z)}$; $\mathbf{X}$ $\sim$ $p_{\theta}\mathbf{(X|z)}$
\end{center}

Due to the intractability of posterior inference of latent variables, the parameter estimation of such graphical model is challenging. This has been tackled by using stochastic gradient variational bayes (SGVB) \cite{kingma2013auto} framework. Here, the proposal distribution $q_{\phi}\mathbf{(z|X)}$ is used to approximate the true posterior  $p\mathbf{(z|X)}$ using variational inference. Combining this inference network $q_{\phi}\mathbf{(z|X)}$ with a generative network $p_{\theta}\mathbf{(X|z)}$ would resemble autoencoding architecture and thus has been known as variational autoencoder. As such, the objective of VAE is to maximize the following variational lower bound with respect to parameters $\theta$ and $\phi$:

\begin{equation}
\begin{aligned}
  \label{eq:VAEObj}
  \sum_{n=1}^{N} \log {p_{\theta}\mathbf{(X_{n})}} \geq \sum_{n=1}^{N} \mathbf{\E}_{q_{\phi}\mathbf{(z_{n}|X_{n})}} [\log {p_{\theta}\mathbf{(X_{n}|z_{n})}}] \\
  - KL[q_{\phi}\mathbf{(z_{n}|X_{n})}||p\mathbf{(z_{n})}] \\
\end{aligned}
\end{equation}

This objective function (\ref{eq:VAEObj}) can be interpreted as minimizing a reconstruction error in the first term along with minimizing the KL-Divergence between the variational approximation of the posterior $q_{\phi}\mathbf{(z|X)}$ and the prior $p\mathbf{(z)}$. The inference network $q_{\phi}\mathbf{(z|X)}$ and generative network $p_{\theta}\mathbf{(X|z)}$ are usually modeled with a neural network.

\subsection{MMD}
Maximum mean discrepancy (MMD) is a closed-form nonparametric two-sample test proposed by \cite{gretton2007kernel}. The purpose of the test involved in MMD is to determine if two samples are from different distributions. If we have two sets of samples $\mathbf{X} = \{\mathbf{x}_{i}\}_{i=1}^{M}$ and $\mathbf{Y} = \{\mathbf{y}_{j}\}_{j=1}^{N}$, then with the use of MMD we are able to determine if generating distribution $P_{X}$ = $P_{Y}$. This is achieved by comparing the statistics between the two samples. A simple test could be the distance between empirical statistics $\phi(.)$ of the two samples:

\begin{equation}
\begin{aligned}
  \label{eq:TestStatistics}
 \norm[\bigg]{\frac{1}{M} \sum_{i=1}^{M} \phi (\mathbf{x}_{i}) - \frac{1}{N} \sum_{j=1}^{N} \phi (\mathbf{y}_{i})}^{2}\\
\end{aligned}
\end{equation}

MMD \cite{gretton2007kernel} is an estimator by expanding the above square such that kernel tricks can be applied. 
\begin{equation}
\begin{aligned}
  \label{eq:MMDKernel}
  MMD[\mathbf{X},\mathbf{Y}] = \frac{1}{M} \sum_{i=1}^{N} k(\mathbf{x}_{i}, \mathbf{x}_{i}) \\- \frac{2}{MN} \sum_{i,j=1}^{M,N} k(\mathbf{x}_{i},\mathbf{y}_{i}) +  \frac{1}{N} \sum_{j=1}^{N} k(\mathbf{y}_{i}, \mathbf{y}_{i})\\
\end{aligned}
\end{equation}

The choice of $\phi$  in (\ref{eq:TestStatistics}) determines what order of statistics we want to match. For instance, a simple identity function leads to matching the sample mean. When kernel trick is applied, the sample vectors are mapped into an infinite dimensional feature space and when this feature space corresponds to universal reproducing kernel hilbert space (RKHS), it is shown in \cite{gretton2012kernel} that asymptotically, MMD is $0$ if and only if $P_{X}$ = $P_{Y}$. Therefore, minimizing MMD for a universal kernel like Gaussian kernel $k(\mathbf{x}, \mathbf{y})$ = $\mathbf{e^{-\beta||\mathbf{x}-\mathbf{y}||^{2}}}$ in (\ref{eq:MMDKernel}) can be interpreted as matching or minimizing certain distance between weighted sums of all raw moments \cite{li2015generative} of $P_{X}$ and $P_{Y}$.

\section{Model}
\subsection{Conditional generative model}
In this work, we introduce a conditional probabilistic model admitting two sources of variations: VT factor $\mathbf{v}$ to capture the general biophysical relationship between origins of ventricular activation and ECG data, and a latent representation $\mathbf{z}$ to capture subject-specific level of adjustments to the data involving factors such as heart and torso anatomy and characteristics of structural abnormality of the heart that are not explicitly observed. First, considering the task-specific factor $\mathbf{v}$ as an observed variable, we frame the conditional generative process as:
\begin{center}
$\mathbf{z}$ $\sim$ $p_{\theta}\mathbf{(z)}$; $\mathbf{X}$ $\sim$ $p_{\theta}\mathbf{(X|z,v)}$
\end{center}

\begin{figure*}[t]
\centering
\includegraphics[width=\linewidth]{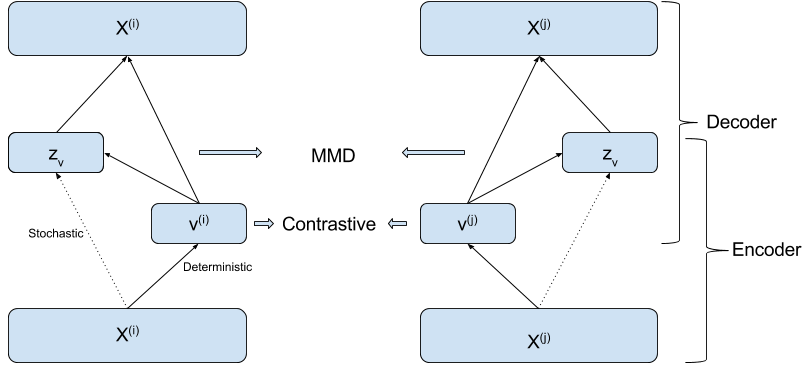}
\caption{\small{Illustrative diagram of the proposed network. The encoder takes a pair of input ECG data and outputs the VT factor $\mathbf{v}$ and subject-specific factor $\mathbf{z}$. The decoder takes these disentangled factors as an input and reconstruct the ECG data.}}
\label{fig:architecture}
\end{figure*}

We model this conditional generative process in the variational autoencoder framework, where $p_{\theta}\mathbf{(X|z,v)}$ is a likelihood function described by a decoder network. The approximate posterior to infer subject-specific factor $\mathbf{z}$ is modeled using an independent Gaussian distribution $q_{\phi}\mathbf{(z|X,v)}$ = $N(\mu, \sigma I)$. This is specified by an encoder network. The stochasticity in $q_{\phi}\mathbf{(z|X,v)}$ comes from both the data-distribution and the randomness of Gaussian distribution. Finally by choosing standard isotropic Gaussian prior $p\mathbf{(z)}$ = $N(0, I)$, we can obtain the following lower bound:

\begin{equation}
\begin{aligned}
  \label{eq:condVAEObj}
  \sum_{n=1}^{N} \log {p_{\theta}\mathbf{(X_{n}|v_{n})}} &\geq \sum_{n=1}^{N} \mathbf{\E}_{q_{\phi}\mathbf{(z_{n}|X_{n},v_{n})}} [\log p_{\theta}\mathbf{(X_{n}|z_{n},v_{n})}] \\
  &\qquad - KL[q_{\phi}\mathbf{(z_{n}|X_{n},v_{n})}||p\mathbf{(z_{n})}] \\
  &= \mathcal{F}(\phi, \theta; \mathbf{X}_{n}, \mathbf{v}_{n}) \\  
\end{aligned}
\end{equation}

The objective in ~\ref{eq:condVAEObj} gives us a simple conditional VAE similar to that used in \cite{DBLP:journals/corr/LouizosSLWZ15}. However, unlike in  \cite{DBLP:journals/corr/LouizosSLWZ15}, neither $\mathbf{v}$ nor $\mathbf{z}$ in this work is observed. Therefore, in this work, $\mathbf{v}$ in  (\ref{eq:condVAEObj}) is obtained using a separate deterministic encoder $q\mathbf{(v|X)}$ in which the only source of stochasticity comes from the data distribution.  As such, the conditional likelihood in  (\ref{eq:condVAEObj}) can be expressed as:

\begin{center}
$\mathbf{X}$ $\sim$ $p_{\theta}(\mathbf{X|z},q\mathbf{(v|X))}$
\end{center}

Although the above model admits two separate sources of causal factors generating the ECG data, there is nothing preventing all of the information from flowing through the latent factor $\mathbf{z}$. In other words, the decoder could learn to ignore the VT factor $\mathbf{v}$. Below, we describe two regularization techniques added to the network to encourage the separation of the two factors and to learn a more discriminative task-specific representation. 

\subsection{Discriminative regularization}
\subsubsection{MMD\\}
The work in \cite{li2015generative} and \cite{dziugaite2015training} used MMD measure to generate samples from an unknown distribution given i.i.d data, essentially, acting as a substitute of difficult adversarial training \cite{goodfellow2014generative}. The idea in \cite{li2015generative} and \cite{dziugaite2015training} is to force the output of the generator network to match all orders of statistics of the training data. Motivated by this idea, we propose to match all order moments between marginal posterior distributions of subject-specific factors $\mathbf{z}$ which is conditioned upon VT factor $\mathbf{v}$ representing different segments in the heart:

\begin{equation}
\begin{aligned}
  \label{eq:MMDObj}
 MMD[\mathbf{z_{v_{i}}},\mathbf{z_{v_{j}}}] &= \norm[\bigg]{ \mathbf{\E}_{p\mathbf{(X|v_{i})}} [\mathbf{\E}_{q\mathbf{(z|X,v_{i})}}[\phi\mathbf{(z_{v_{i}})}]] \\
 &\qquad- \mathbf{\E}_{p\mathbf{(X|v_{j})}} [\mathbf{\E}_{q\mathbf{(z|X,v_{j})}}[\phi\mathbf{(z_{v_{j}})}]]}^{2}\\
\end{aligned}
\end{equation}

\begin{flushleft}
where $i \neq j$  and $k$ = $\{1,..,i,..,j,..,c\}$ represents $c$ categories represented by specific factor $\mathbf{v}$.
\end{flushleft}

Minimizing (\ref{eq:MMDObj}) can be interpreted as forcing the network to learn subject specific factor $\mathbf{z}$ independent of VT factor $\mathbf{v}$. The MMD objective shown in (\ref{eq:MMDObj}) is differentiable when the kernel is differentiable. As such, we are using Gaussian kernels $k(\mathbf{x}, \mathbf{y})$ = $\mathbf{e^{-\beta||\mathbf{x}-\mathbf{y}||^{2}}}$ whose gradient has a simple form and thus could be easily back-propagated through the inference network $q_{\phi}\mathbf{(z|X,v)}$. Bandwidth parameter $\beta$ of kernel $k(.)$ is considered as the hyper-parameter during training of the network.

\subsubsection{Contrastive loss\\}
MMD regularizer would be enough if the VT factor $\mathbf{v}$ is observed and the objective is to learn the subject-specific variations, similar to \cite{DBLP:journals/corr/LouizosSLWZ15}  while learning fair representations. However, in this work, we are also interested in learning representation $\mathbf{v}$ informative for the clinical task at hand. To do so,  we hypothesize that for the ECG data originating from nearby locations, the VT factor $\mathbf{v}$ should be similar regardless if the data are collected from the same patient; otherwise $\mathbf{v}$ should be different. In this way we are learning an embedding of VT factor $\mathbf{v}$ where similar data pairs are embedded nearby and dissimilar data pairs are far apart. To facilitate such pair-wise comparison, we propose an additional contrastive loss \cite{hadsell2006dimensionality} as a weak supervision to the deterministic encoder. The contrastive loss is formulated as:

\begin{equation}
\begin{aligned}
\label{eq:ContrastiveObj}
L_c(\mathbf{v}^{(i)},\mathbf{v}^{(j)}) = & e_v^p\frac{1}{2}\|\mathbf{v}^{(i)} - \mathbf{v}^{(j)}\|_2^2\\
& +(1-e_v^p)\frac{1}{2} \max(0,\alpha-\|\mathbf{v}^{(i)} - \mathbf{v}^{(j)}\|_2^2)
\end{aligned}
\end{equation}

\begin{flushleft}
where $e_v^p$ is 1 if the pair of the ECG data share the same VT exit label and 0 otherwise. Here $\alpha$ is the margin and the dissimilar pairs contribute to the loss function only if their distance is within this margin. We treat $\alpha$ as the hyper-parameter during training of the network. 
\end{flushleft}

\begin{figure*}[t]
\centering
\includegraphics[width=\linewidth]{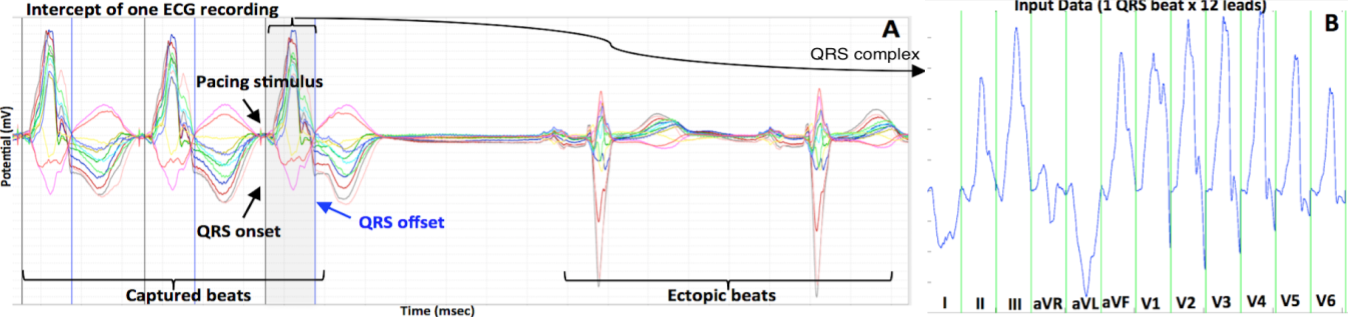}
\caption{\small{(A) Illustration of experimental data and processing. ECG recordings are pre-processed for extraction of noise-free QRS complex. (B) The final data which is formed by concatenating the extracted QRS complex from each lead.}}
\label{fig:Data}
\end{figure*}

To calculate both MMD and contrastive loss, we require a training architecture (presented in next section) which facilitate pair-wise comparison as shown in Fig. ~\ref{fig:architecture}. Combining (\ref{eq:condVAEObj}), (\ref{eq:MMDObj}) and (\ref{eq:ContrastiveObj}), we get our combined objective as:
\begin{equation}
\begin{aligned}
  \label{eq:totalVAEObj}
\mathcal{F}(\phi, \theta, \mathbf{X}_{n}, e_v^p) &=  \sum_{m_{i}, m_{j}}^{M} 
\Big\{ \mathcal{F}(\phi, \theta; \mathbf{X}_{m_{i}}, \mathbf{v}_{m_{i}}) \\
&\qquad + \mathcal{F}(\phi, \theta; \mathbf{X}_{m_{j}}, \mathbf{v}_{m_{j}}) \Big\}/ 2  \\
&\qquad - \lambda_{1} \sum_{m_{i}, m_{j}; m_{i} \neq m_{j}}^{M}MMD[\mathbf{z_{v_{m_{i}}}},\mathbf{z_{v_{m_{j}}}}] \\&\qquad -\lambda_{2}\sum_{m_{i}, m_{j}}^{M} L_c(\mathbf{v}^{(i)},\mathbf{v}^{(j)})
\end{aligned}
\end{equation}

\begin{flushleft}
where $\lambda_{1}$ and $\lambda_{2}$ are the non-negative weight. Here, the upper bound $M$ in the summation refers to the total number of pairs presented to the networks as the input data.  
\end{flushleft}

\subsection{Training architecture}
We randomly generate $M$ training pairs $X^p=(\mathbf{X}^{(i)},\mathbf{X}^{(j)})$ 
from the training data ensuring different beats from the same pacing location are not paired together. Each pair is given a label $e_v^p$ which is 1 if the pair of the ECG data share the same VT exit label and 0 otherwise. The architecture is inspired from \textit{siamese} architecture \cite{chopra2005learning} which consists of two copies of our proposed model sharing the same set of parameters. The input to the entire system is a pair of ECG data ($\mathbf{X}^{(i)},\mathbf{X}^{(j)}$) and a label $e_v^p$. The output of the entire encoder can be represented as \{($\mathbf{v}^{(i)},\mathbf{v}^{(j)}$), ($\mathbf{z}_{v_{i}}, \mathbf{z}_{v_{j}}$)\} which is fed to the decoder network to calculate approximation of the ECG data as ($\mathbf{\tilde{X}}^{(i)},\mathbf{\tilde{X}}^{(j)}$). The illustrative diagram of the network is shown in Fig. \ref{fig:architecture}.

\section{Experiments}
We apply the proposed model to disentangle subject-specific variations present in the ECG data to learn the representation informative for the clinical task of localizing origin of ventricular tachycardia. Following the setup in \cite{yokokawa2012automated}, we divide the left-ventricular surface into ten segments. In this setting, the VT localization process becomes a 10-class classification problem.

\subsection{The Data}
Data are collected from 
39 patients during routine pace-mapping procedures as the patients underwent ablation of scar-related VT. 
The database 
includes 15-second 12-lead ECG recordings 
recorded from 1012 distinctive pacing sites 
on the left-ventricular (LV) endocardium, 
all identified on an intro-operative electroanatomical mapping system (CARTO3) with known coordinates. 
All ECG data are processed with noise removal and baseline correction using an open-source software.
As illustrated in Fig.~\ref{fig:Data}, 
manual selection and extraction of QRS complexes are carried out by 
student trainees 
to avoid motion artifacts, ectopic beats, and non-capture beats.
The final input signal is in the form of one QRS beat from 12 leads, 
down-sampled in time to $100 \times 12$ in size. 
Because multiple quality beats can be extracted from each ECG recording, 
we obtain in-total 16848 sets of ECG data and the corresponding sites of pacing in the form of a segment label. Using these data, we try to learn to predict the origin of the ventricular activation from the ECG data, where the origin is in the form of one of the ten LV segments. When applied to VT, the predicted origin of ventricular activation corresponds to the site where the VT circuit exits the scar.

\subsection{Experimental setup}
To evaluate the proposed model, the entire dataset is split into training, validation and test set as follows: 10292 from 22 patients, 3017 from 5 patients, and 3539 from 12 patients. 
Note that the patients in the test and validation sets are not included in the training set. For each encoder, we have two hidden layers of 800 and 600 units. Similarly for the decoder, we have two hidden layers  of size 600 and 800 units. Different sets of hyperparameters including the size of VT factor $\mathbf{v}$, the size of subject-specific factor $\mathbf{z}$, the margin for contrastive regularization $\alpha$ and the bandwidth parameter $\beta$ of the MMD are tuned on validation set. The scaling parameter $\lambda_{1}$ and $\lambda_{2}$ are set to 1. Rectified linear units (ReLU) were used for the non-linear activation in both encoder and decoder network. The training data is divided into mini-batches and to normalize the data in each mini-batch, batch normalization \cite{ioffe2015batch} layers are used in both encoder and decoder network. The optimization of the objective function was done with stochastic optimizer Adam \cite{kingma2014adam}. Finally the classification task of the learned factor was performed using a simple linear classifier to ensure classifier itself does not aid in boosting the classification performance. The codes associated with this work are available at \url{https://github.com/Prasanna1991/MMD_VAE}. 

\subsection{Results}
We evaluate the performance of our model by first performing classification by associating the learned VT factor into different segments in the left-ventricular region of the heart.  We also perform experiments to evaluate and analyze the disentanglement abilities of the network. 

\subsubsection{Classification \\}
The results for classification of VT factor $\mathbf{v}$ into pre-defined ten segments for proposed model along with other different models are presented in Table ~\ref{tab:result}. The localization accuracy in terms of classification accuracy of the proposed model is compared against commonly used prescribed features of 120-ms QRS Integral \cite{sapp2012automated} , standard VAE without disentangling the two representations and finally with standard VAE with MMD regularization where disentanglement between subject-specific factor and task-specific VT factor is achieved by matching all order of moments between the marginal posterior distributions of subject-specific factor conditioned on task-specific VT factors representing different ventricular segments of the heart. To be consistent with the network architecture, all models are implemented with the same \textit{siamese} architecture. The results are presented in Table ~\ref{tab:result} where we can see an improvement as we go from feature based approach to deep network in the form of standard VAE. Further improvement is observed with VAE+MMD model where model is encouraged to separate VT factor $\mathbf{v}$ with subject-specific factor $\mathbf{z}$. Finally, classification is further improved with the proposed model where contrastive regularizer is added for the learning of VT factor $\mathbf{v}$ on top of VAE+MMD model. 

\begin{table}[t]
  \centering
  \begin{tabular}[t]{c|c}
    \hline
    Architecture & Segment classification \\
    & (in \%)\\
    \hline
    QRS Integral & 45.28 \\ 
    VAE & 52.37 \\
    VAE+MMD & 53.17 \\
    \textbf{Proposed} & \textbf{55.29}\\
    \hline 
  \end{tabular} 
  \vspace{.1cm}
  \caption{Segment classification accuracy of the proposed method versus three comparison methods. Classification accuracy is reported in percentage on separately held out test dataset.}
  \label{tab:result}
\end{table}

\begin{figure*}[h]
\centering
\includegraphics[width=\linewidth, height=8cm]{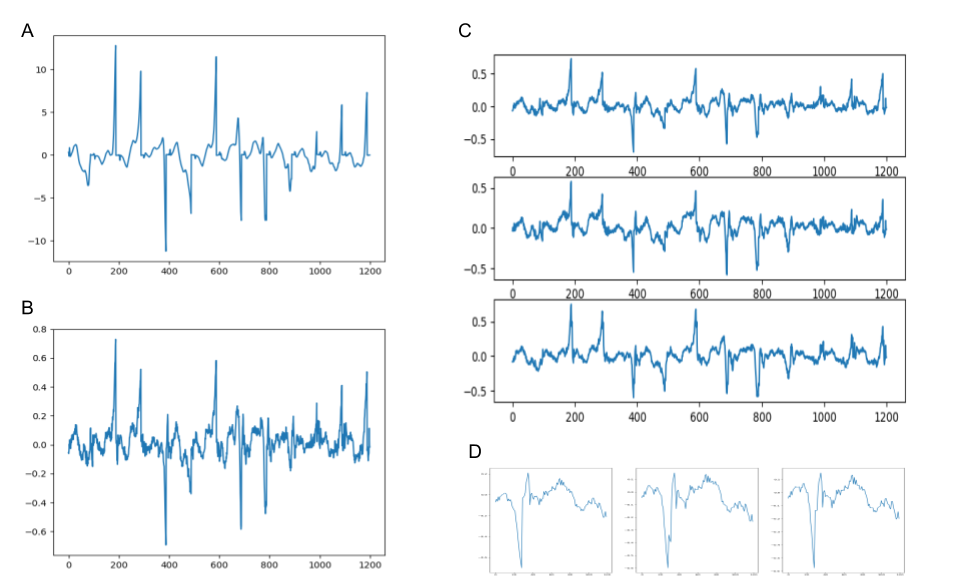}
\caption{\small{(A) Original ECG as presented to the model. (B) Reconstructed ECG. (C) Generated ECG signal by setting $\mathbf{z}$ = $N(0,I)$ and $\mathbf{v}$ from different ECG signals with the same VT labels. (D) Lead aVR of the generated ECG as shown in (C) for a more detailed view.}}
\label{fig:ECG}
\end{figure*}

\subsubsection{Analyzing disentanglement \\}
To gain an insight into the effect of factor disentangling, we performed different experiments on both causal factors to the ECG data. First, we examine if information of one factor is removed from the representation learned for other factor, i.e., whether information about VT origin is removed from the patient-specific variation, and vice versa. To do so, we attempt to use the learned representation for inter-subject variations $\mathbf{z}$ to classify the location of the VT, and use the learned representation for VT location $\mathbf{v}$ to recognize the patient ID. Both of them are compared with random chance as shown in Table ~\ref{tab:resultDis}. As already seen in Table ~\ref{tab:result}, VT factor $\mathbf{v}$ is informative for the specific task of localizing the origin of ventricular activation. In comparison, Table ~\ref{tab:resultDis} demonstrates that the subject-specific factor $\mathbf{z}$ has minimal information about the specific-task. On the other hand, subject-specific factor $\mathbf{z}$ is much better in associating with different patient ID present in training data compared to VT factor $\mathbf{v}$.

Further, we present visualization of ECG signals used and generated from the experiments in Fig. ~\ref{fig:ECG}. For the purpose of analyzing disentanglement we generate different ECG signals by setting $z$ = $N(0,I)$ as shown in Fig. ~\ref{fig:ECG} (C). For VT factor $\mathbf{v}$ we use the representation learned from $q(\mathbf{v}|\mathbf{X})$ where $\mathbf{X}$ represents different ECG signals with the same VT label. Although the ECG data comes from different subjects, because of the same VT factor, the generated ECG signal looks alike in all three cases. However, there still exist variations and to demonstrate such variations, we present the detailed view by selecting one of the lead (lead aVR) among 12-leads in Fig. ~\ref{fig:ECG} (D) where maximum variations can be observed.     
\begin{table}[t]
  \centering
  \begin{tabular}[t]{c|cc}
    \hline
    factor & VT location & patient ID  \\
    & (in \%) & (in \%) \\
    \hline
    $\mathbf{v}$ & 55.29 & 15.23  \\ 
    \hline 
    $\mathbf{z}$ & 17.19 & 32.33 \\ 
    \hline
    random-chance & 10 & 4.5 \\
    \hline 
  \end{tabular} 
  \vspace{.1cm}
  \caption{Classification accuracy when one factor is associated with label of the other factor. Classification accuracy involving subject-specific factor $\mathbf{z}$ is reported on train dataset.}
  \label{tab:resultDis}
\end{table}

\section{Discussion}
The results presented in this work are obtained when the models only see 22 patients during training, a very limited sample size for the purpose of disentangling inter-subject variations. Moreover, as the data were collected from actual pace-mapping procedures during ablation, available data on each patient typically covers only a specific region of the heart that is within and around the myocardial scar. The networks within the models are constructed with a simple multilayer perceptron (MLP) and using powerful feature extractor like convolutional neural network (CNN) \cite{DBLP:journals/corr/RajpurkarHHBN17} within the encoder network and/or the decoder network may further improve the performance. The ability to separate the subject-specific factor with the factor informative for clinical task at hand, at the presence of a limited dataset, however, does demonstrate the feasibility of the proposed model.

\section{Conclusion}
The paper investigates the first use of deep generative models to disentangle inter-subject variations within the ECG data from the representation informative for the clinical task at hand. The feasibility of the proposed framework is demonstrated by a novel yet important clinical task to localize the VT exit which may enhance the efficiency and efficacy of ablation therapies. The paper presents a conditional variational autoencoder to extract the subject-specific adjustments to the ECG data, conditioned on task-specific representations learned from a deterministic encoder. To encourage independence between these representations, kernel based MMD measure is used. The learning of task-specific representations from a deterministic encoder is performed using weakly supervised contrastive regularization. 

\section{Acknowledgement}
This work is supported in part by  the National Institutes of Health [No: R21HL125998] and the National Science Foundation [No: ACI-1350374

\bibliographystyle{aaai}
\bibliography{clinical}

\begin{thebibliography}{}

\bibitem[\protect\citeauthoryear{Bengio, Courville, and
  Vincent}{2013}]{bengio2013representation}
Bengio, Y.; Courville, A.; and Vincent, P.
\newblock 2013.
\newblock Representation learning: A review and new perspectives.
\newblock {\em IEEE transactions on pattern analysis and machine intelligence}
  35(8):1798--1828.

\bibitem[\protect\citeauthoryear{Chen \bgroup et al\mbox.\egroup
  }{2016}]{chen2016infogan}
Chen, X.; Duan, Y.; Houthooft, R.; Schulman, J.; Sutskever, I.; and Abbeel, P.
\newblock 2016.
\newblock Infogan: Interpretable representation learning by information
  maximizing generative adversarial nets.
\newblock In {\em Advances in Neural Information Processing Systems},
  2172--2180.

\bibitem[\protect\citeauthoryear{Chen \bgroup et al\mbox.\egroup
  }{2017}]{chen2017}
Chen, S.; Gyawali, P.~K.; Liu, H.; Horacek, B.~M.; Sapp, J.~L.; and Wang, L.
\newblock 2017.
\newblock Disentangling inter-subject variations: Automatic localization of
  ventricular tachycardia origin from 12-lead electrocardiograms.
\newblock  616--619.
\newblock IEEE.

\bibitem[\protect\citeauthoryear{Chopra, Hadsell, and
  LeCun}{2005}]{chopra2005learning}
Chopra, S.; Hadsell, R.; and LeCun, Y.
\newblock 2005.
\newblock Learning a similarity metric discriminatively, with application to
  face verification.
\newblock In {\em Computer Vision and Pattern Recognition, 2005. CVPR 2005.
  IEEE Computer Society Conference on}, volume~1,  539--546.
\newblock IEEE.

\bibitem[\protect\citeauthoryear{Desjardins, Courville, and
  Bengio}{2012}]{desjardins2012disentangling}
Desjardins, G.; Courville, A.; and Bengio, Y.
\newblock 2012.
\newblock Disentangling factors of variation via generative entangling.
\newblock {\em arXiv preprint arXiv:1210.5474}.

\bibitem[\protect\citeauthoryear{Dziugaite, Roy, and
  Ghahramani}{2015}]{dziugaite2015training}
Dziugaite, G.~K.; Roy, D.~M.; and Ghahramani, Z.
\newblock 2015.
\newblock Training generative neural networks via maximum mean discrepancy
  optimization.
\newblock {\em arXiv preprint arXiv:1505.03906}.

\bibitem[\protect\citeauthoryear{Goodfellow \bgroup et al\mbox.\egroup
  }{2014}]{goodfellow2014generative}
Goodfellow, I.; Pouget-Abadie, J.; Mirza, M.; Xu, B.; Warde-Farley, D.; Ozair,
  S.; Courville, A.; and Bengio, Y.
\newblock 2014.
\newblock Generative adversarial nets.
\newblock In {\em Advances in neural information processing systems},
  2672--2680.

\bibitem[\protect\citeauthoryear{Gretton \bgroup et al\mbox.\egroup
  }{2007}]{gretton2007kernel}
Gretton, A.; Borgwardt, K.~M.; Rasch, M.; Sch{\"o}lkopf, B.; and Smola, A.~J.
\newblock 2007.
\newblock A kernel method for the two-sample-problem.
\newblock In {\em Advances in neural information processing systems},
  513--520.

\bibitem[\protect\citeauthoryear{Gretton \bgroup et al\mbox.\egroup
  }{2012}]{gretton2012kernel}
Gretton, A.; Borgwardt, K.~M.; Rasch, M.~J.; Sch{\"o}lkopf, B.; and Smola, A.
\newblock 2012.
\newblock A kernel two-sample test.
\newblock {\em Journal of Machine Learning Research} 13(Mar):723--773.

\bibitem[\protect\citeauthoryear{Hadsell, Chopra, and
  LeCun}{2006}]{hadsell2006dimensionality}
Hadsell, R.; Chopra, S.; and LeCun, Y.
\newblock 2006.
\newblock Dimensionality reduction by learning an invariant mapping.
\newblock In {\em Computer vision and pattern recognition, 2006 IEEE computer
  society conference on}, volume~2,  1735--1742.
\newblock IEEE.

\bibitem[\protect\citeauthoryear{Higgins \bgroup et al\mbox.\egroup
  }{2016}]{higgins2016early}
Higgins, I.; Matthey, L.; Glorot, X.; Pal, A.; Uria, B.; Blundell, C.; Mohamed,
  S.; and Lerchner, A.
\newblock 2016.
\newblock Early visual concept learning with unsupervised deep learning.
\newblock {\em arXiv preprint arXiv:1606.05579}.

\bibitem[\protect\citeauthoryear{Hu \bgroup et al\mbox.\egroup
  }{2017}]{hu2017toward}
Hu, Z.; Yang, Z.; Liang, X.; Salakhutdinov, R.; and Xing, E.~P.
\newblock 2017.
\newblock Toward controlled generation of text.
\newblock In {\em International Conference on Machine Learning},  1587--1596.

\bibitem[\protect\citeauthoryear{Ioffe and Szegedy}{2015}]{ioffe2015batch}
Ioffe, S., and Szegedy, C.
\newblock 2015.
\newblock Batch normalization: Accelerating deep network training by reducing
  internal covariate shift.
\newblock In {\em International Conference on Machine Learning},  448--456.

\bibitem[\protect\citeauthoryear{Kingma and Ba}{2014}]{kingma2014adam}
Kingma, D., and Ba, J.
\newblock 2014.
\newblock Adam: A method for stochastic optimization.
\newblock {\em arXiv preprint arXiv:1412.6980}.

\bibitem[\protect\citeauthoryear{Kingma and Welling}{2013}]{kingma2013auto}
Kingma, D.~P., and Welling, M.
\newblock 2013.
\newblock Auto-encoding variational bayes.
\newblock In {\em Proceedings of the 2nd International Conference on Learning
  Representations (ICLR)}, number 2014.

\bibitem[\protect\citeauthoryear{Kingma \bgroup et al\mbox.\egroup
  }{2014}]{kingma2014semi}
Kingma, D.~P.; Mohamed, S.; Rezende, D.~J.; and Welling, M.
\newblock 2014.
\newblock Semi-supervised learning with deep generative models.
\newblock In {\em Advances in Neural Information Processing Systems},
  3581--3589.

\bibitem[\protect\citeauthoryear{Kulkarni \bgroup et al\mbox.\egroup
  }{2015}]{kulkarni2015deep}
Kulkarni, T.~D.; Whitney, W.~F.; Kohli, P.; and Tenenbaum, J.
\newblock 2015.
\newblock Deep convolutional inverse graphics network.
\newblock In {\em Advances in Neural Information Processing Systems},
  2539--2547.

\bibitem[\protect\citeauthoryear{Li, Swersky, and
  Zemel}{2015}]{li2015generative}
Li, Y.; Swersky, K.; and Zemel, R.
\newblock 2015.
\newblock Generative moment matching networks.
\newblock In {\em Proceedings of the 32nd International Conference on Machine
  Learning (ICML-15)},  1718--1727.

\bibitem[\protect\citeauthoryear{Louizos \bgroup et al\mbox.\egroup
  }{2015}]{DBLP:journals/corr/LouizosSLWZ15}
Louizos, C.; Swersky, K.; Li, Y.; Welling, M.; and Zemel, R.~S.
\newblock 2015.
\newblock The variational fair autoencoder.
\newblock {\em CoRR} abs/1511.00830.

\bibitem[\protect\citeauthoryear{Mathieu \bgroup et al\mbox.\egroup
  }{2016}]{mathieu2016disentangling}
Mathieu, M.~F.; Zhao, J.~J.; Zhao, J.; Ramesh, A.; Sprechmann, P.; and LeCun,
  Y.
\newblock 2016.
\newblock Disentangling factors of variation in deep representation using
  adversarial training.
\newblock In {\em Advances in Neural Information Processing Systems},
  5040--5048.

\bibitem[\protect\citeauthoryear{Park, Kim, and
  Marchlinski}{2012}]{park2012using}
Park, K.-M.; Kim, Y.-H.; and Marchlinski, F.~E.
\newblock 2012.
\newblock Using the surface electrocardiogram to localize the origin of
  idiopathic ventricular tachycardia.
\newblock {\em Pacing and Clinical Electrophysiology} 35(12):1516--1527.

\bibitem[\protect\citeauthoryear{Plonsey and Fleming}{1969}]{plonsey}
Plonsey, R., and Fleming, D.~G.
\newblock 1969.
\newblock {\em Bioelectric phenomena}.
\newblock New York: McGraw-Hill.

\bibitem[\protect\citeauthoryear{Rajpurkar \bgroup et al\mbox.\egroup
  }{2017}]{DBLP:journals/corr/RajpurkarHHBN17}
Rajpurkar, P.; Hannun, A.~Y.; Haghpanahi, M.; Bourn, C.; and Ng, A.~Y.
\newblock 2017.
\newblock Cardiologist-level arrhythmia detection with convolutional neural
  networks.
\newblock {\em CoRR} abs/1707.01836.

\bibitem[\protect\citeauthoryear{Reed \bgroup et al\mbox.\egroup
  }{2014}]{reed2014learning}
Reed, S.; Sohn, K.; Zhang, Y.; and Lee, H.
\newblock 2014.
\newblock Learning to disentangle factors of variation with manifold
  interaction.
\newblock In {\em Proceedings of the 31st International Conference on Machine
  Learning (ICML-14)},  1431--1439.

\bibitem[\protect\citeauthoryear{Rezende, Mohamed, and
  Wierstra}{2014}]{DBLP:conf/icml/RezendeMW14}
Rezende, D.~J.; Mohamed, S.; and Wierstra, D.
\newblock 2014.
\newblock Stochastic backpropagation and approximate inference in deep
  generative models.
\newblock In {\em Proceedings of the 31th International Conference on Machine
  Learning, {ICML} 2014, Beijing, China, 21-26 June 2014},  1278--1286.

\bibitem[\protect\citeauthoryear{Sapp \bgroup et al\mbox.\egroup
  }{2012}]{sapp2012automated}
Sapp, J.~L.; El-Damaty, A.; MacInnis, P.~J.; Warren, J.~W.; and
  Hor{\'a}{\v{c}}ek, B.~M.
\newblock 2012.
\newblock Automated localization of pacing sites in postinfarction patients
  from the 12-lead electrocardiogram and body-surface potential maps.
\newblock {\em Computing in Cardiology}.

\bibitem[\protect\citeauthoryear{Stevenson}{2009}]{stevenson2009ventricular}
Stevenson, W.~G.
\newblock 2009.
\newblock Ventricular scars and ventricular tachycardia.
\newblock {\em Transactions of the American Clinical and Climatological
  Association} 120:403.

\bibitem[\protect\citeauthoryear{Tenenbaum and
  Freeman}{2000}]{tenenbaum2000separating}
Tenenbaum, J.~B., and Freeman, W.~T.
\newblock 2000.
\newblock Separating style and content with bilinear models.
\newblock {\em Neural computation} 12(6):1247--1283.

\bibitem[\protect\citeauthoryear{Wiens and Guttag}{2010}]{wiens2010active}
Wiens, J., and Guttag, J.~V.
\newblock 2010.
\newblock Active learning applied to patient-adaptive heartbeat classification.
\newblock In {\em Advances in neural information processing systems},
  2442--2450.

\bibitem[\protect\citeauthoryear{Yokokawa \bgroup et al\mbox.\egroup
  }{2012}]{yokokawa2012automated}
Yokokawa, M.; Liu, T.-Y.; Yoshida, K.; Scott, C.; Hero, A.; Good, E.; Morady,
  F.; and Bogun, F.
\newblock 2012.
\newblock Automated analysis of the 12-lead electrocardiogram to identify the
  exit site of postinfarction ventricular tachycardia.
\newblock {\em Heart Rhythm} 9(3):330--334.

\end{thebibliography}
\end{document}